# DRL-Based Injection Molding Process Parameter Optimization for Adaptive and Profitable Production

Joon-Young Kim [1,2*], Jecheon Yu [1*], Heekyu Kim [1], Seunghwa Ryu [1‡]

[1] Department of Mechanical Engineering, Korea Advanced Institute of Science and Technology (KAIST), Daejeon, Republic of Korea
[2] Industrial Intelligence Research Group, AI/DX Center, Institute for Advanced Engineering (IAE), Yongin, Republic of Korea

**Abstract:** Plastic injection molding remains essential to modern manufacturing. However, optimizing process parameters to balance product quality and profitability under dynamic environmental and economic conditions remains a persistent challenge. This study presents a novel deep reinforcement learning (DRL)-based framework for real-time process optimization in injection molding, integrating product quality and profitability into the control objective. A profit function was developed to reflect real-world manufacturing costs, incorporating resin, mold wear, and electricity prices, including time-of-use variations. Surrogate models were constructed to predict product quality and cycle time, enabling efficient offline training of DRL agents using soft actor-critic (SAC) and proximal policy optimization (PPO) algorithms. Experimental results demonstrate that the proposed DRL framework can dynamically adapt to seasonal and operational variations, consistently maintaining product quality while maximizing profit. Compared to traditional optimization methods such as genetic algorithms, the DRL models achieved comparable economic performance with up to 135× faster inference speeds, making them well-suited for real-time applications. The framework's scalability and adaptability highlight its potential as a foundation for intelligent, data-driven decision-making in modern manufacturing environments.

**Keywords:** Plastic injection molding; Process parameter optimization; Profit optimization; Adaptive production systems; Deep reinforcement learning (DRL); Proximal policy optimization (PPO); Soft actor-critic (SAC)

# 1 INTRODUCTION

## 1.1 Research Background

Plastic injection molding is a cornerstone of modern manufacturing, enabling the mass production of plastic components with high precision, repeatability, and efficiency [1,2]. Its applications span various industries, including automotive, electronics, medical devices, and consumer goods, where lightweight and durable plastic components have replaced mainly traditional materials such as metal and glass. The capacity to produce complex geometries at scale with minimal material waste has further established injection molding as a vital manufacturing process [3].

Despite widespread adoption, the injection molding industry operates under intense cost pressures due to its high-volume, low-margin nature. Manufacturers must maintain a delicate balance between ensuring product quality and optimizing operational costs [4]. Inefficiencies such as excessive material usage, prolonged cycle times, and elevated electricity consumption can significantly erode profit margins [5–8]. This challenge is exacerbated by external factors, including fluctuations in raw material prices, electricity costs, and seasonal variations, all of which can impact the overall profitability of injection molding operations [9,10].

Traditionally, process parameters in injection molding, such as injection pressure, mold temperature, cooling time, and holding pressure, have been optimized with a primary focus on product quality. Numerous studies have investigated the influence of these parameters on defect rates, mechanical performance, and surface finish [11]. In recent years, statistical models and machine learning techniques have been increasingly adopted to predict optimal process conditions for consistent product quality [12–16].

However, these methodologies fail to address a pivotal element of dynamic cost fluctuations in real-world manufacturing contexts. These methodologies optimize for quality but do not account for the financial impact of changing electricity prices, equipment operation, and maintenance costs.

As a result, manufacturers may achieve high-quality production but fail to maximize profitability, leading to potential financial losses despite optimal technical performance. For example, electricity pricing varies significantly based on time of use, with peak-hour electricity costs often being considerably higher than off-peak rates [17–20]. In regions where electricity pricing follows a dynamic pricing structure, operating injection molding machines during peak hours without adjusting process parameters accordingly can lead to unnecessarily high production costs. Environmental conditions such as ambient temperature and relative humidity can also affect machine performance, influencing cycle times and cooling efficiency [21–26]. When process settings remain static, they cannot respond effectively to these external variations, leading to increased electricity consumption and reduced cost-effectiveness.

To overcome these limitations, developing a decision-making model capable of dynamically adjusting process parameters in response to real-time operational and environmental conditions is essential. This necessitates transitioning from conventional static optimization methods to an adaptive, real-time decision-making approach. Reinforcement learning (RL) methods present a promising solution because they allow systems to learn continuously from operational feedback and to improve decision-making strategies based on changing cost and quality considerations [27–29].

This study proposes a deep reinforcement learning (DRL)-based framework for optimizing injection molding process parameters with the dual objectives of ensuring product quality and maximizing profitability. Traditional optimization approaches have primarily focused on reducing defects under fixed operating conditions, often overlooking the economic impact of dynamic variables such as electricity pricing, resin costs, and mold wear. To address this gap, the proposed framework integrates cost-related factors directly into the optimization process, enabling adaptive control in variable manufacturing environments [30–33].

The framework employs soft actor-critic (SAC) [34,35] and proximal policy optimization (PPO) [34] algorithms, which are well-suited for continuous control tasks with stability and sample efficiency. Surrogate models were developed using production data collected through a Design of Experiments (DOE) methodology to predict product quality and cycle time. These models allow the DRL agent to be trained offline in a simulated environment that closely reflects real-world process dynamics. Once deployed, the agent optimizes process parameters in real time, adjusting to fluctuating environmental and cost conditions within sub-second inference time.

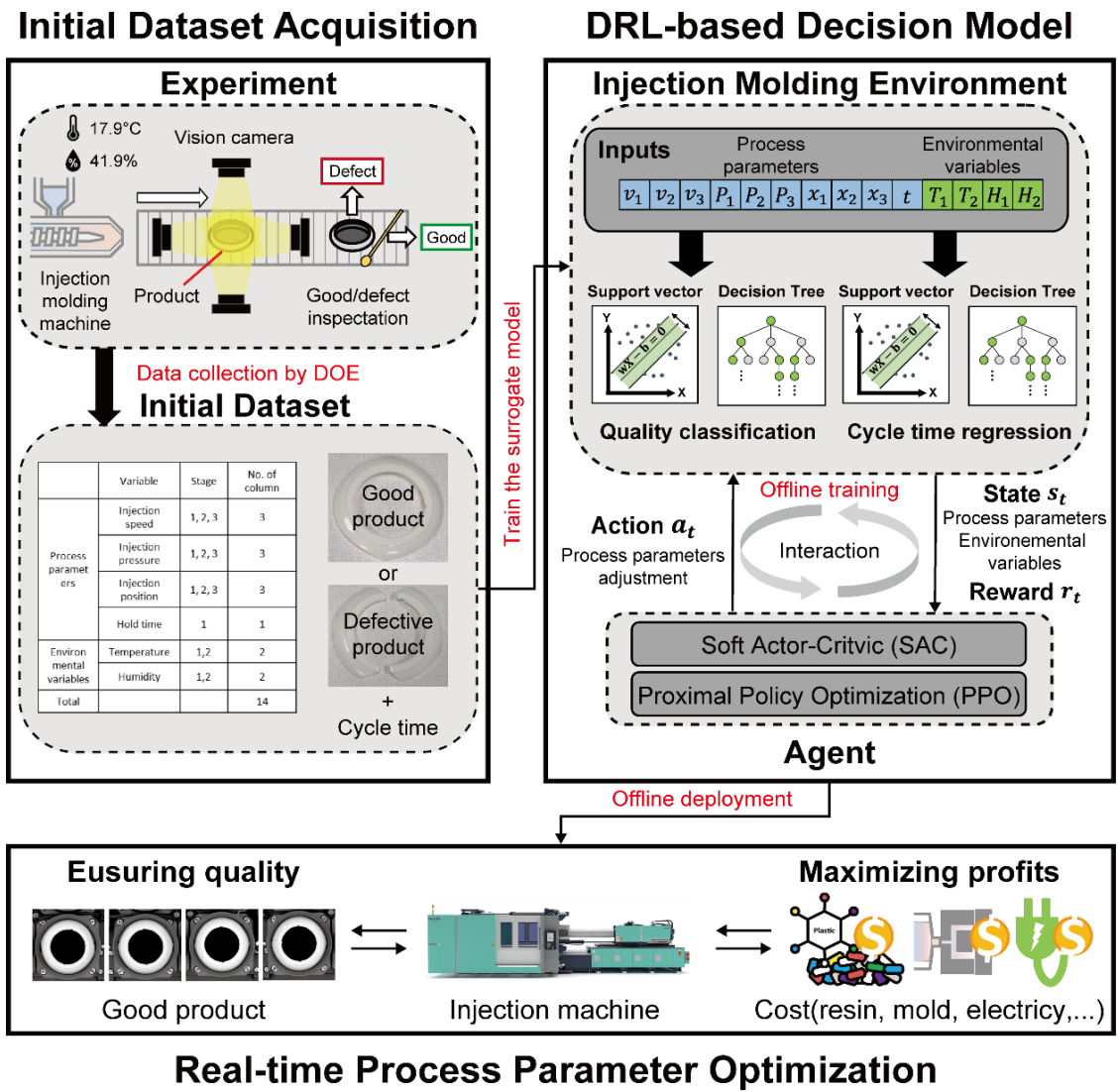

**Fig. 1**. Framework for real-time process parameter optimization using deep reinforcement learning considering environmental variations in injection molding processes.

As illustrated in **Fig. 1**, the proposed framework achieves profitability levels comparable to global optimization methods such as genetic algorithms while offering over 100-fold faster decision-making. This approach not only supports real-time manufacturing control but also contributes to the development of intelligent, cost-aware production systems that are scalable and applicable across various industrial contexts. The results demonstrate the feasibility of DRL as a practical solution for sustainable, economically optimized manufacturing operations.

The structure of this paper is as follows. **Section 1.2** presents a comprehensive review of prior studies on the application of Deep Learning (DL) and RL for injection molding process optimization. **Section 2** outlines the problem definition, incorporating the formulation of the profit function and the Markov Decision Process (MDP) of DRL-based decision-making models. It also introduces the surrogate models used for quality classification and cycle time prediction, as well as the theoretical foundations of PPO and SAC. **Section 3** describes the experimental results, including data acquisition, surrogate model evaluation, and DRL agent for process parameter optimization. This section also provides a comparative analysis of PPO and SAC, benchmarked against a Genetic Algorithm (GA), and presents deployment results of the DRL agents under three different seasonal scenarios in a virtual production environment. **Section 4** discusses the implications, limitations, and future research directions of the proposed DRL framework. Finally, **Section 5** summarizes key findings and highlights their implications for the injection molding industry.

## 1.2 Related works

Injection molding is a complex manufacturing process that requires precise control over parameters such as injection pressure, mold temperature, cooling time, and holding pressure to ensure both product quality and operational efficiency. Traditional optimization approaches, including DOE and Response Surface Methodology (RSM), have been extensively employed to identify optimal process settings and improve performance. DOE has been successfully applied in optimizing injection molding for various applications such as automotive components, micro-injection processes, and geometrically complex parts [34–36]. These studies have demonstrated improvements in defect reduction, dimensional accuracy, and part stability. DOE has also been used to minimize contour distortion in composite components and to enhance structural integrity in molded products [37]. A combined simulation and experimental approach has validated DOE-driven parameter selection for thin-shell plastic parts, ensuring dimensional consistency and defect reduction [38]. RSM, particularly when integrated with advanced optimization techniques, has facilitated fine-tuning of process parameters [39]. For instance, combining RSM with nonlinear programming and Artificial Neural Networks (ANNs) has enabled the precise optimization of molding processes for plastic lenses and reinforced polycarbonate composites, resulting in better mechanical properties and reduced variability [40,41]. However, these methods generally require extensive experimental trials and may be limited in their ability to capture the nonlinear and dynamic nature of modern injection molding systems [42,43].

With the advancement of computational intelligence, machine learning (ML) and deep learning (DL) techniques have emerged as powerful tools for modeling complex interactions between process parameters and product outcomes. One study applied various DOE methods with artificial neural networks (ANN) to assess their impact on modeling efficiency and accuracy [44]. Another developed a regression-based ML system for real-time defect prediction, improving production efficiency and reducing waste [45]. ML has also been used to reveal links between processing conditions, structural features, and material properties in polypropylene molding, enabling parameter optimization for better mechanical performance [46]. In addition, supervised learning has been employed to predict cooling time, aiding in production planning and cycle time reduction [47]. DL techniques have addressed class imbalance in fault detection, enhancing defect classification and process control [48], while DL-based optimal tracking control methods have been proposed to monitor resin flow in molding machines [49]. These studies highlight the growing role of deep learning in optimizing injection molding, improving accuracy in defect detection, parameter prediction, and process stability, ultimately leading to better product quality and manufacturing efficiency.

RL is a method in which an agent interacts with an environment to learn an optimal policy for maximizing rewards without requiring expert knowledge, making it an attractive approach for various applications. Similarly, DRL has emerged by combining RL with deep neural networks, allowing agents to learn effectively in broader and more complex state and action spaces [30,35]. Advances in DRL have gained significant attention by demonstrating impressive performance in highly complex games such as Go [36] and StarCraft II [37]. Recently, DRL has also been actively applied in the manufacturing domain. For instance, its applications have been reported in areas such as system design [38–40], process planning [41,42] process control and optimization [43–46] and quality control [47,48].

In the domain of injection molding, RL has been applied to real-time process parameter optimization. One study proposed a decision-making framework that used offline data and a self-predictive ANN to dynamically adjust parameters and maintain consistent product quality [50]. However, this approach relied on simulation-based surrogate models, which may introduce modeling inaccuracies, and utilized the Deep Deterministic Policy Gradient (DDPG) algorithm, which is known to suffer from instability and limited exploration in complex, high-dimensional spaces. Moreover, most existing studies have primarily focused on quality objectives, such as dimensional accuracy or part thickness, without integrating cost or profit metrics into the learning process. Other studies have explored various RL strategies for optimizing molding processes. RL has been employed to refine ram velocity and packing pressure profiles [27], optimize flow distribution in liquid composite molding [49], and improve trajectory control using Iterative Learning Control (ILC) combined with RL [50]. Approaches such as Deep Q-Networks (DQN) have been applied to achieve intelligent decision-making under complex production conditions [30]. In addition, DRL has been utilized for intelligent temperature control in stretch blow molding and for temperature compensation using Actor-Critic framework, surpassing traditional PID and GPC controllers in stability and accuracy [32,51]. Further developments include RL-based self-recovery mechanisms for correcting non-optimal conditions [52], two-dimensional Q-learning for unknown system tracking [53], and fault-tolerant tracking control using off-policy RL for robust operation under actuator failures [54].

In summary, while deep learning and RL have significantly advanced the optimization of injection molding processes, integrating profitability into these models remains an emerging area. It is imperative to acknowledge the limitations of previous studies and to develop more accurate surrogate models, employ robust RL algorithms such as PPO or SAC, and expand the scope of the considered process parameters. This approach will lead to more adaptive and profitable production strategies.

## 2 METHODOLOGY

### 2.1 Data Acquisition

A real-world dataset was employed to develop the DRL agent capable of optimizing process parameters in injection molding. The data were acquired from a fully instrumented injection molding testbed, which has been described in detail in our prior study [55]. The experimental setup included the installation of environmental sensors to monitor factory and machine-level temperature and relative humidity, integrating a Central Monitoring System (CMS) for capturing process parameters, and deploying an automated vision inspection system for product quality assessment.

The collected dataset comprises process parameters such as injection speed, pressure, position, and hold time, along with external environmental variables including temperature and relative humidity, both within the factory and at the machine level. A total of 2,794 samples were collected. Each data sample comprises 10 controllable process parameters, 4 external environmental variables, and 1 binary quality label (good or defective), which was automatically assigned through visual inspection. The target product for the experiment was a circular cosmetic container cap made from acrylonitrile butadiene styrene (ABS), a thermoplastic material commonly used in industrial applications.

**Table 1.** Statistical analysis results of the dataset.

|  | Variable name | Mean | Std Dev | Min | 25% | 50% | 75% | Max | Unit |
|---|---|---|---|---|---|---|---|---|---|
| Process parameter | Injection speed 1 | 29.5 | 6.5 | 20.0 | 25.0 | 30.0 | 35.0 | 40.0 | % |
|  | Injection speed 2 | 30.1 | 6.5 | 20.0 | 25.0 | 30.0 | 35.0 | 40.0 | % |
|  | Injection speed 3 | 20.1 | 6.5 | 10.0 | 15.0 | 20.0 | 25.0 | 30.0 | % |
|  | Injection pressure 1 | 129.9 | 6.5 | 120.0 | 125.0 | 130.0 | 135.0 | 140.0 | bar |
|  | Injection pressure 2 | 130.1 | 6.5 | 120.0 | 125.0 | 130.0 | 135.0 | 140.0 | bar |
|  | Injection pressure 3 | 140.1 | 6.5 | 130.0 | 135.0 | 140.0 | 145.0 | 150.0 | bar |
|  | Injection position 1 | 46.0 | 1.3 | 44.0 | 45.0 | 46.0 | 47.0 | 48.0 | mm |
|  | Injection position 2 | 37.8 | 3.9 | 32.0 | 35.0 | 38.0 | 41.0 | 44.0 | mm |
|  | Injection position 3 | 30.0 | 1.3 | 28.0 | 29.0 | 30.0 | 31.0 | 32.0 | mm |
|  | Hold time | 1.2 | 0.8 | 0.0 | 0.6 | 1.2 | 1.8 | 2.4 | sec |
| Environmental variables | Machine temperature | 15.4 | 4.2 | 5.5 | 13.3 | 14.5 | 19.8 | 21.8 | °C |
|  | Machine humidity | 42.1 | 9.8 | 23.6 | 33.0 | 44.7 | 51.0 | 62.4 | % |
|  | Factory temperature | 15.3 | 4.4 | 5.6 | 12.1 | 14.3 | 19.9 | 22.8 | °C |
|  | Factory humidity | 42.6 | 10.6 | 23.0 | 33.2 | 47.1 | 50.8 | 63.6 | % |

The environmental conditions were recorded continuously throughout the production process and naturally fluctuated over time. Conversely, process parameters were systematically varied according to a DOE scheme, specifically utilizing an L81 orthogonal array. This methodological design ensured broad parameter space coverage and improved the dataset's representativeness concerning the complex, nonlinear interactions present in real-world injection molding environments. For a detailed description of the experimental configuration, the testbed architecture, and quality labeling methodology, readers are directed to our previous work. A statistical summary of the dataset is presented in **Table 1**.

## 2.2 Problem definition and MDP formulation of DRL-based decision-making model

The primary objective of this research is to develop a DRL-based optimization framework that maximizes profitability in the injection molding process. Conventional approaches have primarily focused on optimizing process parameters to enhance product quality, often without incorporating critical economic factors such as electricity consumption, mold costs, and overall production efficiency. To address this limitation, a profit function is formulated that integrates both quality-related and cost-related components. This formulation supports a holistic and financially sustainable optimization strategy.

The profit function, which represents profit per a single production cycle, is formulated as follows:

$$f(x) = p \sum_{i=1}^{Cv} y_i - \sum_{i=1}^{Cv} \left( c_i^{\text{resin}} + c_i^{\text{mold}}(P_{\max}) + c_i^{\text{elect}}(P_{\max}) \right) \quad (1)$$

Where,

- $Cv$ represents the number of cavities per cycle (in this case, $Cv = 4$).
- $p$ represents the unit price per cavity (in this case, $p = 0.2\$/\text{cavity}$).
- $y_i$ represents a binary variable that indicates whether the product from the $i$-th cavity is good (1) or defective (0).

$$y_i = \begin{cases} 1, & \text{if } i\text{- th cavity is good cavity} \\ 0, & \text{if } i\text{- th cavity is defective cavity} \end{cases} \quad (2)$$

- $c_i^{\text{resin}}$ represents the resin cost per cavity (in this case, $c_i^{\text{resin}} = 0.04\$/\text{cavity}$).
- $c_i^{\text{mold}}(P_{\max})$ represents the mold cost per cavity ($/cavity), which is determined based on the maximum injection pressure. This cost is derived from a method utilized by the factory in this study for cost analysis, where the estimated mold wears per cycle and remaining usable cycles are calculated based on maximum injection pressure. The cost

per cavity is obtained by distributing the total mold replacement cost over 4 cavities per cycle.

$$c_i^{\text{mold}}(P_{\max}) = \begin{cases} 0.025, & \text{if } P_{\max} < 140 \text{ bar} \\ 0.02775, & \text{if } 140 \text{ bar} \leq P_{\max} \end{cases} \quad (3)$$

- $c_i^{\text{elect}}(P_{\max})$ represents the electricity cost per cavity ($/cavity), which depends on the maximum injection pressure. The electricity consumption per cycle is 0.8 kWh for $P_{\max} < 135$ bar, 1.0 kWh for $135 \text{ bar} \leq P_{\max} < 145 \text{ bar}$, and 1.1 kWh for $145 \text{ bar} \leq P_{\max}$. Dividing by 4 cavities per cycle gives per-cavity values of 0.2 kWh, 0.25 kWh, and 0.275 kWh. The electricity cost per cavity is then determined by applying seasonal and time-dependent electricity pricing. **Tables 2** and **3** summarize the time-of-use electricity prices and their seasonal classifications. **Fig. 2** illustrates the corresponding hourly price variations across seasons.

$$c_i^{\text{elect}}(P_{\max}) = \begin{cases} 0.2 \times \text{electricity price}, & \text{if } P_{\max} < 135 \text{ bar} \\ 0.25 \times \text{electricity price}, & \text{if } 135 \text{ bar} \leq P_{\max} < 145 \text{ bar} \\ 0.275 \times \text{electricity price}, & \text{if } 145 \text{ bar} \leq P_{\max} \end{cases} \quad (4)$$

By maximizing the profit function, the DRL agent can learn an optimal control policy that balances product quality, operational efficiency, and cost minimization in dynamic production environments. The proposed DRL-based framework is designed to adjust process parameters autonomously in real time. This allows it to respond effectively to fluctuating external conditions, thereby enabling adaptive and data-driven decision-making. Several key factors are considered in the problem definition. Environmental conditions, including ambient temperature and humidity, significantly influence resin behavior and affect flow dynamics, necessitating real-time parameter adjustments to maintain acceptable product quality. Another essential factor is cycle time, which must be minimized to enhance productivity while still ensuring that products meet quality standards. In addition, electricity cost variability plays a role, as electricity prices fluctuate depending on time and season, requiring optimized scheduling to reduce costs. The mold cost is also crucial because frequent high-pressure cycles accelerate wear, shortening mold lifespan. Therefore, the agent must balance production speed with the long-term health of the mold.

**Table 2.** Time-of-use electricity prices in South Korea in USD/kWh.
(The exchange rate is 1,000 KRW to a USD)

|                | Spring/Fall | Summer | Winter |
|----------------|-------------|--------|--------|
| Off-peak hours | 0.0995      | 0.0995 | 0.1065 |
| Mid-peak hours | 0.1220      | 0.1524 | 0.1526 |
| On-peak hours  | 0.1527      | 0.2345 | 0.2101 |

**Table 3.** Time-of-use classification by season in South Korea.

|                | Spring/Fall                                           | Summer                                                | Winter                                                |
|----------------|-------------------------------------------------------|-------------------------------------------------------|-------------------------------------------------------|
| Off-peak hours | 22:00 - 08:00                                         | 22:00 - 08:00                                         | 22:00 - 08:00                                         |
| Mid-peak hours | 08:00 - 11:00<br>12:00 - 13:00<br>18:00 - 22:00       | 08:00 - 11:00<br>12:00 - 13:00<br>18:00 - 22:00       | 08:00 - 09:00<br>12:00 - 16:00<br>19:00 - 22:00       |
| On-peak hours  | 11:00 - 12:00<br>13:00 - 18:00                        | 11:00 - 12:00<br>13:00 - 18:00                        | 09:00 - 12:00<br>16:00 - 19:00                        |

**Table 4.** Ranges of process parameters including upper-lower bounds, small-large step size.

| Process parameter    | Lower bound | Upper bound | Small step size | Large step size |
|----------------------|-------------|-------------|-----------------|-----------------|
| Injection speed 1    | 20          | 40          | 0.1             | 5.0             |
| Injection speed 2    | 20          | 40          | 0.1             | 5.0             |
| Injection speed 3    | 10          | 30          | 0.1             | 5.0             |
| Injection pressure 1 | 120         | 140         | 1.0             | 20              |
| Injection pressure 2 | 120         | 140         | 1.0             | 20              |
| Injection pressure 3 | 130         | 150         | 1.0             | 20              |
| Injection position 1 | 44          | 48          | 0.1             | 1.0             |
| Injection position 2 | 32          | 44          | 0.1             | 1.0             |
| Injection position 3 | 28          | 32          | 0.1             | 1.0             |
| Hold time            | 0           | 2.4         | 0.1             | 1.0             |

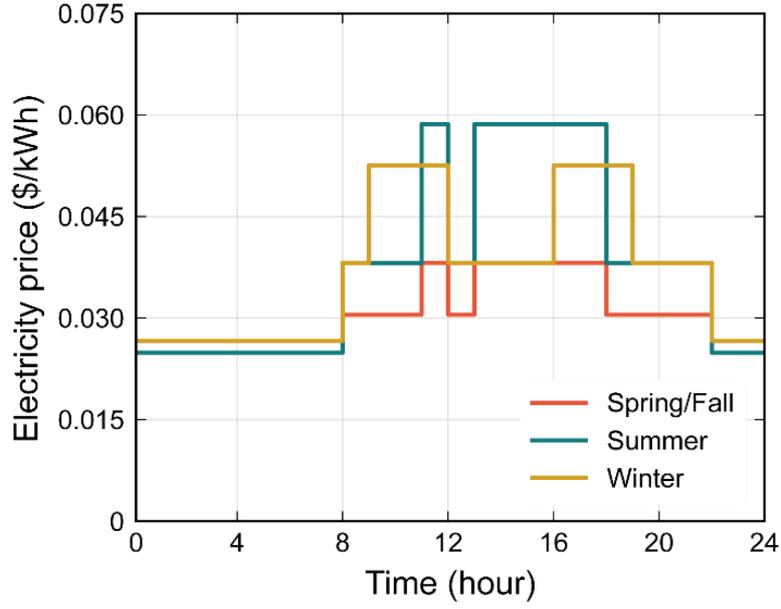

**Fig. 2**. Seasonal time-of-use electricity price profiles by hour for spring/fall, summer, and winter.

To enable the DRL agent to make optimal decisions in dynamic production environments, the decision-making problem is formulated as MDP. The MDP framework provides a mathematical foundation for sequential decision-making, aligning well with the challenges of injection molding, where external conditions and operational parameters continuously evolve. An MDP is characterized by a tuple $(S, A, R, T)$, representing the state space, action space, reward function, and transition probability, respectively [56,57]. This study defines these components to reflect the specific requirements of injection molding optimization.

First, a state space $S$ represents the set of all possible states that the environment can assume. A state $s_t \in S$ is defined as a vector that incorporates not only the controllable process parameters but also external environmental variables and electricity price variations, all of which critically affect product quality, cycle time, and production costs:

$$s_t = [p_t, e_t, c_t] \in S \tag{5}$$

where $p_t$ denotes the vector of controllable process parameters, $e_t$ represents the environmental variables, and $c_t$ corresponds to the electricity pricing information at time $t$.

The process parameter vector $p_t$ is defined as:

$$p_t = [v_t^1, v_t^2, v_t^3, P_t^1, P_t^2, P_t^3, x_t^1, x_t^2, x_t^3, h_t] \quad (6)$$

where $v_t^k$, $P_t^k$, and $x_t^k$ represent the injection speed, pressure, and position at three distinct stages ($k = 1, 2, 3$) of the injection process, and $h_t$ denotes hold time. The environmental variable vector $e_t$ is defined as:

$$e_t = [T_t^{\text{machine}}, T_t^{\text{factory}}, H_t^{\text{machine}}, H_t^{\text{factory}}] \quad (7)$$

representing the temperature and relative humidity conditions at both the machine and factory. The electricity price $c_t$ is encoded as a 9-dimensional one-hot vector, reflecting seasonal and time-dependent price categories outlined in **Tables 2** and **3**. To stabilize the learning process, the continuous elements of the state vector are normalized to the range $[-1, 1]$, using the upper and lower bounds listed in **Table 4**.

Second, an action space $A$ denotes the set of possible adjustments that the agent can make to the process parameters. An action $a_t \in A$ is defined as a continuous adjustment vector, allowing the agent to control the process parameters under varying environmental conditions:

$$a_t = [\Delta v_t^1, \Delta v_t^2, \Delta v_t^3, \Delta P_t^1, \Delta P_t^2, \Delta P_t^3, \Delta x_t^1, \Delta x_t^2, \Delta x_t^3, \Delta h_t] \in A \quad (8)$$

where each $\Delta$ component represents the adjustment to the corresponding controllable process parameter: injection speed $(v_t^k)$, pressure $(P_t^k)$, position $(x_t^k)$, and hold time $(h_t)$ at different stages $(k = 1, 2, 3)$. The evolution of process parameters over time follows a cumulative adjustment mechanism formulated as:

$$p_t = p_0 + \sum_{i=0}^{t-1} a_i \quad (9)$$

where the current process parameters $p_t$ are determined by adding the initial process parameter setting $p_0$ and the cumulative sum of the past adjustment actions up to timestep $t - 1$. To stabilize the learning process and maintain consistency with the normalized state representation, the DRL agent outputs actions normalized within the range $[-1, 1]$. In the environment, these normalized actions are rescaled using predefined adjustment step sizes specific to each process parameter, as listed in **Table 5**. Two types of step sizes are defined: (i) a small step size, corresponding to one fourth of the difference between the upper and lower bounds of each process parameters, and (ii) a larger step size, corresponding to one half of that difference. Furthermore, to ensure operational feasibility, any action resulting in a combination of process parameters exceeding its allowed bounds, as specified in **Table 4**, is clipped accordingly.

**Table 5** Adjustment step sizes for normalized actions used in the environment.

| Process parameter | Small step size | Large step size |
| --- | --- | --- |
| Injection speed 1 | 5.0 | 10.0 |
| Injection speed 2 | 5.0 | 10.0 |
| Injection speed 3 | 5.0 | 10.0 |
| Injection pressure 1 | 5.0 | 10.0 |
| Injection pressure 2 | 5.0 | 10.0 |
| Injection pressure 3 | 5.0 | 10.0 |
| Injection position 1 | 1.0 | 2.0 |
| Injection position 2 | 3.0 | 6.0 |
| Injection position 3 | 1.0 | 2.0 |
| Hold time | 0.6 | 1.2 |

Third, the reward function $R$ assigns a reward $r_t = R(s_t, a_t)$ based on the action $a_t$ taken in state $s_t$. In the defined problem, the agent generates newly adjusted process parameters, which are then applied to the injection molding machine. After executing the injection process with these parameters, the reward is computed as the profit over a 10-minute production interval, formulated based on the profit function defined in **Eq. 1** and expressed as:

$$r_t = \frac{600}{T} \times \left[ p \sum_{i=1}^{C_v} y_i - \sum_{i=1}^{C_v} \left( c_i^{\text{resin}} + c_i^{\text{mold}}(P_{\max}) + c_i^{\text{elect}}(P_{\max}) \right) \right] \quad (10)$$

where $T$ represents the cycle time in seconds, $C_v$ is the number of cavities per cycle, $y_i$ indicates the quality of i-th cavity. $c_i^{\text{resin}}$, $c_i^{\text{mold}}$, and $c_i^{\text{elect}}$ denote the resin cost, mold cost, and electricity cost per cavity within a single production cycle, respectively. The choice to formulate the reward based on a 10-minute interval is made to match the setting where each timestep within an episode corresponds to a 10-minute production period during training phase. Both the cycle time ($T$) and the cavity quality labels ($y_i$) are functions of the process parameters and environmental variables. These quantities are inferred during training using the trained surrogate models embedded within the environment, enabling rapid reward computation without real-world experiments. The surrogate models are described in detail in **Section 2.3**.

Fourth, the transition probability $P$ represents the likelihood of transitioning to the next state $s_{t+1}$ when an action $a_t$ is taken in the current state $s_t$, expressed as $p(s_{t+1}|s_t, a_t)$. It defines the state transition dynamics within the MDP framework. The agent does not explicitly model or estimate this transition function in model-free RL. Instead, it learns through trial-and-error interactions with the environment, implicitly capturing the underlying transition dynamics from sampled transitions. Despite the absence of an explicit transition model, agents can successfully learn state value functions, state-action value functions, and optimal policies by leveraging temporal difference methods [56].

### 2.3 Surrogate models for quality classification and cycle time regression

A reliable simulation environment was required to enable efficient offline training of the DRL agent. This was achieved by developing two surrogate models. One model was designed to predict product quality, while the other was developed to estimate cycle time. These models serve as key components of the virtual environment where the DRL agent can simulate the outcomes of different process parameters without directly interacting with the physical injection molding machine.

The first surrogate model performs quality classification by predicting whether a product manufactured under a given set of process parameters is acceptable or defective. This model was initially developed and validated in our previous study on diffusion-based quality estimation and process parameter inference in real manufacturing environments [55]. It was trained using production data labeled with quality outcomes, along with controllable process parameters and environmental variables. The model replicates quality behavior observed in the real-world testbed and outputs a binary label representing the predicted quality classification.

In contrast, the surrogate model for cycle time regression was newly developed in this study to support the novel objective of profitability-oriented process optimization. This regression model estimates production cycle time based on given process parameters. To identify the most effective regression algorithm, 18 ML models were tested using the PyCaret package [58], including Random Forest, Gradient Boosting, Light Gradient Boosting Machine (LightGBM), Decision Tree, and others. The dataset was split into training and testing sets in an 80:20 ratio, with 10% of the training data further allocated to a validation set. Model evaluation was conducted through 10-fold cross-validation, and the average performance scores across folds were compared. Although RMSE (Root Mean Squared Error) and MAE (Mean Absolute Error) are widely used metrics in regression evaluation, this study adopted $R^2$ (coefficient of determination) as the performance metric. RMSE can disproportionately penalize outliers due to the squaring of errors. While MAE offers a straightforward average error, it does not capture the model's ability to explain the variance of the target variable. Both are also scale-dependent and thus less suitable for comparing models across varying process settings. In contrast, $R^2$ offers a normalized, interpretable measure of how much variance in the dependent variable is accounted for by the model, making it more suitable for evaluating predictive performance in complex, multivariable environments like injection molding. Using 10-fold cross-validation, LightGBM was identified as the best-performing model among the 18 tested, achieving the highest average $R^2$ score. Its excellent predictive accuracy, fast training time, and strong generalization capability made it an ideal choice for modeling the nonlinear and dynamic nature of cycle time in real manufacturing scenarios. The results are presented in **Table 6**. Further details of surrogate models are provided in **Section 1** of the **Supplementary Materials**.

**Table 6.** The performance results from 10-fold cross-validation on 18 ML models for cycle time regression.

| Model | RMSE | MAE | $R^2$ |
| --- | --- | --- | --- |
| Light Gradient Boosting Machine | 0.1468 | 0.0632 | 0.9743 |
| Random Forest Regressor | 0.1537 | 0.0583 | 0.9719 |
| Extra Trees Regressor | 0.1559 | 0.0451 | 0.9710 |
| Decision Tree Regressor | 0.1674 | 0.0475 | 0.9666 |
| Gradient Boosting Regressor | 0.1749 | 0.0986 | 0.9645 |
| K Neighbors Regressor | 0.3776 | 0.2285 | 0.8365 |
| AdaBoost Regressor | 0.3916 | 0.3123 | 0.8246 |
| Ridge Regression | 0.7239 | 0.6270 | 0.4011 |
| Bayesian Ridge | 0.7239 | 0.6271 | 0.4011 |
| Linear Regression | 0.7239 | 0.6270 | 0.4011 |
| Elastic Net | 0.9142 | 0.7613 | 0.0467 |
| Orthogonal Matching Pursuit | 0.9333 | 0.7768 | 0.0064 |
| Lasso Regression | 0.9388 | 0.7830 | -0.0052 |
| Lasso Least Angle Regression | 0.9388 | 0.7830 | -0.0052 |
| Dummy Regressor | 0.9391 | 0.7832 | -0.0058 |
| Huber Regressor | 1.1059 | 0.8215 | -0.4008 |
| Least Angle Regression | 1.1678 | 0.7724 | -0.5819 |
| Passive Aggressive Regressor | 1.3215 | 1.0765 | -1.0907 |

## 2.4 DRL algorithms

Model-free DRL algorithms can be categorized into value-based and policy-based methods. In value-based methods, the agent first learns a value function and subsequently derives a policy based on it. However, value-based approaches are generally restricted to discrete action spaces and are not readily applicable to continuous action domains. To overcome this limitation, policy-based methods learn a policy directly, enabling the handling of continuous action spaces [59]. Nevertheless, policy-based approaches often suffer from high gradient variance, negatively impacting learning stability and overall performance. A framework that integrates both approaches is the actor-critic method, wherein the actor is responsible for learning the policy, and the critic estimates the value function to guide and stabilize policy updates [60].

Additionally, DRL algorithms can be classified into on-policy and off-policy methods depending on whether the policy used for training matches the policy that determines the actions. In on-policy methods, the policy being updated is identical to the one that determined the action. By contrast, off-policy methods enable learning from samples collected by a different policy. Consequently, on-policy methods cannot reuse samples generated by previous policies and must rely on newly collected trajectories for each training iteration. In contrast, off-policy methods can leverage past experiences by storing them in an experience replay buffer and incorporating them into training. On-policy methods offer more stable learning and exhibit better convergence properties. However, they are susceptible to getting trapped in local optima and suffer from high sample complexity. Off-policy methods are potentially less stable but achieve greater sample efficiency by extensively reusing prior experiences [56,61,62].

On-policy DRL algorithms collect data using the current policy and update the policy based on those collected samples. Trusted region policy optimization (TRPO) [63] ensures stable policy improvement by restricting updates within a trust region. However, TRPO suffers from high computational complexity and implementation difficulty. PPO [34] addresses these limitations by introducing a clipped surrogate objective $L^{\text{CLIP}}(\theta)$, as shown in **Eq. 11**, thereby replacing the complex trust region constraint in TRPO. The actor network is trained to maximize this surrogate objective. To further enhance the stability of policy updates, PPO employs Generalized Advantage estimation (GAE) $\hat{A}_t^{\text{GAE}}$, which is defined in **Eq. 12** and **Eq. 13**. The actor network is trained to maximize $L^{\text{actor}}$ defined in **Eq. 14**, which incorporates an entropy bonus term $S[\pi_\theta](s_t)$ to encourage sufficient exploration. The critic network estimates the state value function $V_\phi(s_t)$, which is utilized in computing the GAE and guiding updates of the actor network. The critic is trained by minimizing the loss function $L^{\text{critic}}(\phi)$ given in **Eq. 15**.

$$L^{\text{CLIP}}(\theta) = \hat{E}_t \left[ \frac{\pi_\theta(s_t)}{\pi_{\theta_{old}}(s_t)} \hat{A}_t^{\text{GAE}}, \text{clip}\left( \frac{\pi_\theta(s_t)}{\pi_{\theta_{old}}(s_t)}, 1 - \epsilon, 1 + \epsilon \right) \hat{A}_t^{\text{GAE}} \right] \qquad (11)$$

$$\hat{A}_t = \delta_t + (\gamma\lambda)\delta_{t+1} + \cdots + \cdots + (\gamma\lambda)^{T-t+1} \tag{12}$$

$$where\ \delta_t = r_t + \gamma V_\phi(s_{t+1}) - V_\phi(s_t) \tag{13}$$

$$L^{\text{actor}} = \hat{E}_t[L_t^{\text{CLIP}}(\theta) + cS[\pi_\theta](s_t)] \tag{14}$$

$$L^{\text{critic}}(\phi) = \hat{E}_t\left[\left(V_\phi(s_t) - \hat{R}_t\right)^2\right] \tag{15}$$

where $\theta$ and $\phi$ denote the parameters of the actor and critic network, respectively. The term $\frac{\pi_\theta(s_t)}{\pi_{\theta_{old}}(s_t)}$ represents the probability ratio between the current policy and the previous policy. The hyperparameter $\epsilon, \gamma$, and $\lambda$ represent the clipping parameter, the discount factor, and GAE smoothing parameter, respectively. $\delta_t$ denotes the temporal difference error. $S[\pi_\theta](s_t)$ denotes the entropy of the policy $\pi_\theta$ at state $s_t$, with $c$ as its weighting coefficient. The estimated return $\hat{R}_t$ is computed as the discounted sum of future rewards.

Various off-policy algorithms have been developed to achieve high sample efficiency by reusing samples from experience replay to address challenges associated with continuous state and action spaces. A representative approach is Deep Deterministic Policy Gradient (DDPG) [35], which optimizes a deterministic policy. However, DDPG relies on external noise for exploration rather than incorporating stochasticity into the policy itself, making it sensitive to hyperparameter settings and prone to unstable training. In contrast, SAC [64,65] adopts a stochastic policy and introduces entropy regularization by incorporating the expected policy entropy $H(\pi(s_t))$ in **Eq. 16**. This framework enables the policy to explicitly balance exploration and exploitation, facilitating broad action sampling during the early training phase and mitigating premature convergence to local optima. The policy objective $\hat{J}_\pi(\theta)$ is formulated in **Eq. 17** using the minimum of two soft Q-value functions, implementing a clipped double Q-learning to reduce overestimation bias. The critic consists of four networks: two Q-networks for learning and two corresponding target networks. The Q-networks are trained by minimizing the loss defined in **Eq.**

**18**, where the target value is computed according to **Eq. 19**. The target networks are updated via a soft update as defined in **Eq. 20**.

$$J(\pi) = \sum_{t=0}^{T} E_t[r(s_t, a_t) + \alpha H(\pi(s_t))] \tag{16}$$

$$\hat{J}_\pi(\theta) = \hat{E}_t\left[\alpha \log \log \pi_\theta(s_t) - Q_{\phi_j}(s_t, a_t)\right] \tag{17}$$

$$\hat{J}_Q(\phi_i) = E_t\left[(y - Q_{\phi_i}(s_t, a_t))^2\right] \tag{18}$$

$$y = r_{t+1} + \gamma\left(Q_{\phi_{\text{targ},j}}(s_t, a_t) - \alpha \log \pi_\theta(s_{t+1})\right) \tag{19}$$

$$\phi_{targ,i} = \tau\phi_i + (1-\tau)\phi_{\text{targ},i} \tag{20}$$

where $\theta$, $\phi$, and $\phi_{\text{targ}}$ are parameters of the actor network, the Q-network, and the corresponding target network, respectively. $H(\pi(s_t))$ denotes the entropy of the policy, and $\alpha$ is the entropy temperature coefficient. $Q(s_t, a_t)$ represents the soft Q-value function, which estimates the expected return given state $s_t$ and action $a_t$. $\tau \in (0,1)$ is the Polyak averaging coefficient used for target network updates.

**Table 7**. Summary of the hyperparameters used for the PPO and SAC algorithms

| Hyperparameters | PPO | SAC |
| --- | --- | --- |
| Policy learning rate | $3 \times 10^{-4}$ | $3 \times 10^{-4}$ |
| Value learning rate | $1 \times 10^{-3}$ | $3 \times 10^{-4}$ |
| Total timesteps | 180,000 | 180,000 |
| Buffer size (Experience replay) | 720 | 25,000 |
| Discount factor $\gamma$ | 0.99 | 0.99 |
| Clipping parameter $\epsilon$ | 0.2 | N/A |
| GAE smoothing parameter $\lambda$ | 0.95 | N/A |
| Entropy coefficient $c$ | 0.005 | N/A |
| Learning starts (steps) | N/A | 2016 |
| Target update rate $\tau$ | N/A | 0.005 |
| Training frequency (steps) | N/A | 72 |
| Entropy temperature coefficient $\alpha$ | N/A | 0.1 |
| Hidden layer sizes | [256, 256] | [256, 256] |
| Activation function | Tanh | ReLU |
| Optimizer | Adam; betas=(0.9, 0.999) | Adam; betas=(0.9, 0.999) |

The optimal policy was trained using PPO and SAC with hyperparameters summarized in **Table 7**. Each episode corresponded to a 24-hour period comprising 144 steps with 10-minute intervals. The training process was conducted over 1,250 episodes, resulting in a total of 180,000 steps. All training procedures were executed on an Intel Core i9-13900K CPU, and the DRL framework was implemented based on the OpenAI Gym library [66].

### 2.5 Training and deployment process in DRL

The DRL agent receives a state input that includes the current process parameters and environmental conditions such as the temperature, relative humidity, and electricity prices. Based on this state, the agent selects an action corresponding to adjusting the process parameters, which is applied to the environment. However, conducting real-world experiments to train the agent using process parameters derived from virtual environmental scenarios would be prohibitively expensive. The surrogate models introduced in **Section 2.3** simulate the outcomes of different process settings to address this challenge. These models enable the agent to interact with a virtual environment in an offline setting and learn to optimize the reward function defined in **Eq. 10**. This reward is designed to encourage both profitability and product quality under variable operating conditions.

**Fig. 3** illustrates the framework of the proposed decision-making model for optimizing injection molding process parameters. The framework consists of an offline training phase and an online deployment phase. In the offline training phase, the DRL agent is trained under a virtual environmental scenario rather than using real-time environmental variables from the actual machine. The virtual environmental scenario simulated variations in temperature and relative humidity within the machine and factory over 144 timesteps, corresponding to a total duration of 24 hours, and is treated as one training episode. The magnitude of environmental variations is restricted to 10% of the range between the maximum and minimum environmental values designed based on the DOE as shown in **Table 1**. Additionally, electricity price fluctuations are introduced randomly over time to reflect dynamic cost conditions. The DRL agent is trained across a series of randomly generated virtual environmental scenarios, aiming to develop a policy that maximizes profit under varying conditions and enhances robustness against environmental variability. During training, the agent adopts a stochastic policy modeled by a normal distribution, enabling the sampling of random actions and facilitating effective exploration.

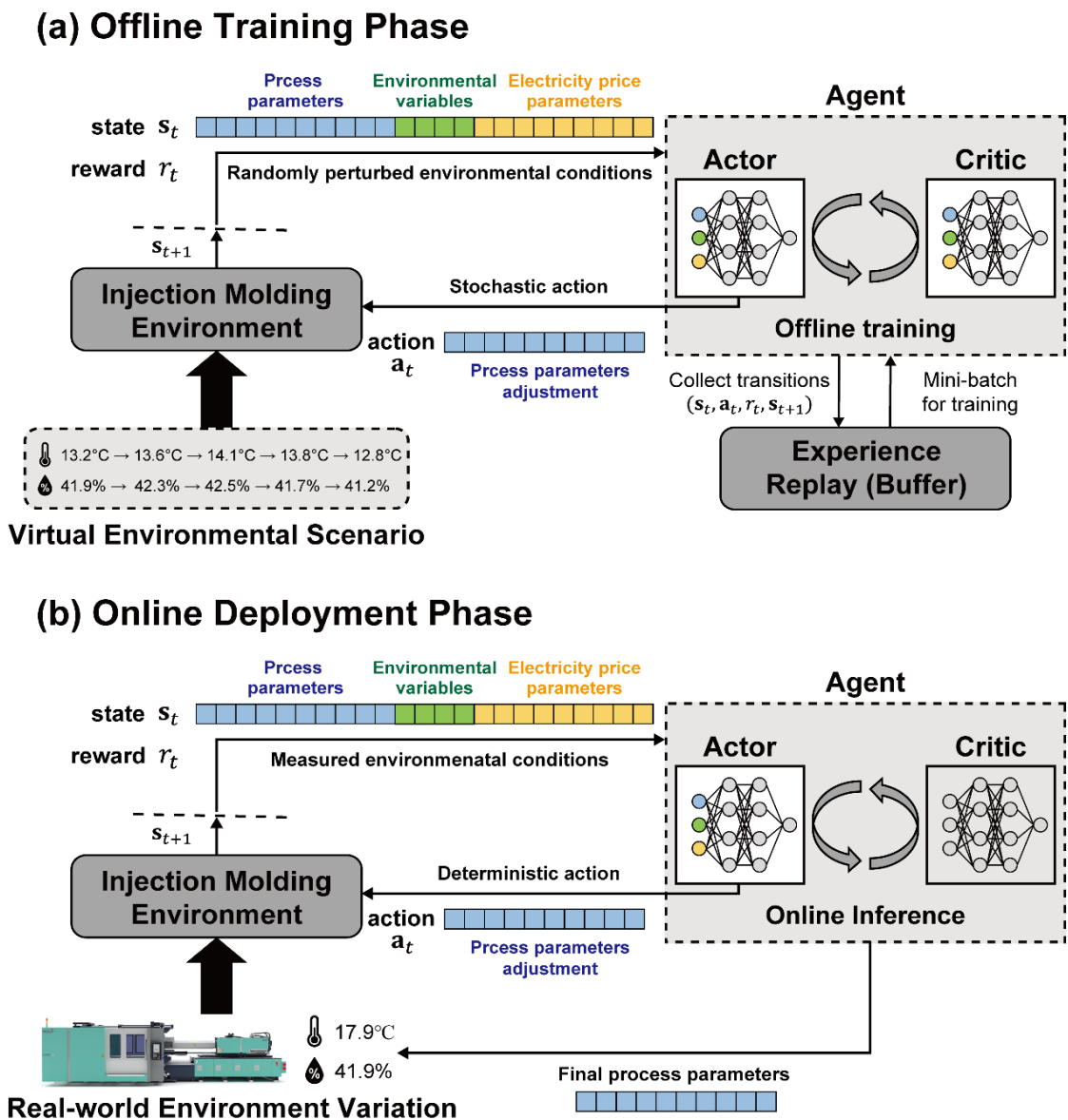

**Fig. 3**. Training phase and deployment phase of the DRL-based decision-making model for injection molding process parameter optimization

In the online deployment phase, the DRL agent operates within a real-world production environment, receiving inputs such as sensor-measured temperature and relative humidity as well as electricity prices that vary by season and time. Leveraging the policy trained during the offline phase, the agent optimizes process parameters within approximately 0.5 seconds under the given environmental conditions. During deployment, a deterministic policy is adopted by selecting the means of the normal distribution, ensuring that identical states consistently yield identical actions. This approach enhances the reliability of the decision-making process. It should be noted that only the actor network is utilized during deployment, whereas both the actor and critic networks are required during training.

## 3  RESULTS

### 3.1  Performance evaluation of the developed DRL-based decision-making models

#### 3.1.1  Training and evaluation results of developed DRL models

In the training phase, each episode corresponds to one of the 1,250 distinct predefined virtual scenarios, and the initial process parameters are randomly initialized to enhance the robustness of the learned policy. A higher reward indicates that the trained policy effectively adjusts the process parameters to maximize profit under given process conditions, environmental variables, and electricity prices.

**Figs. 4a** and **4b** illustrate the change in reward over 180,000 training steps, equivalent to 1,250 episodes, for two DRL algorithms. The curve represents the moving average of 50 episodes, and the shaded region denotes one standard deviation around the mean. Since each episode is associated with a distinct scenario with randomly initialized process parameters, the maximum achievable reward varies across episodes, which leads to a certain degree of variability in the reward curves in the training phase. Nevertheless, both algorithms exhibit convergence of the average reward toward approximately 6.3. In addition, SAC exhibits slightly faster convergence compared to PPO. This is attributed to its higher sample efficiency as an off-policy algorithm, whereas PPO relies on recently collected data for training, necessitating a larger number of samples. **Figs. 4c** and **4d** show the number of defective cavities produced during training, approaching zero as the learning process. However, some defects still occur even after the policy has sufficiently converged, particularly when the given initial process parameters inherently lead to the production of defective cavities. SAC also exhibits a faster reduction in defective cavities compared to PPO.

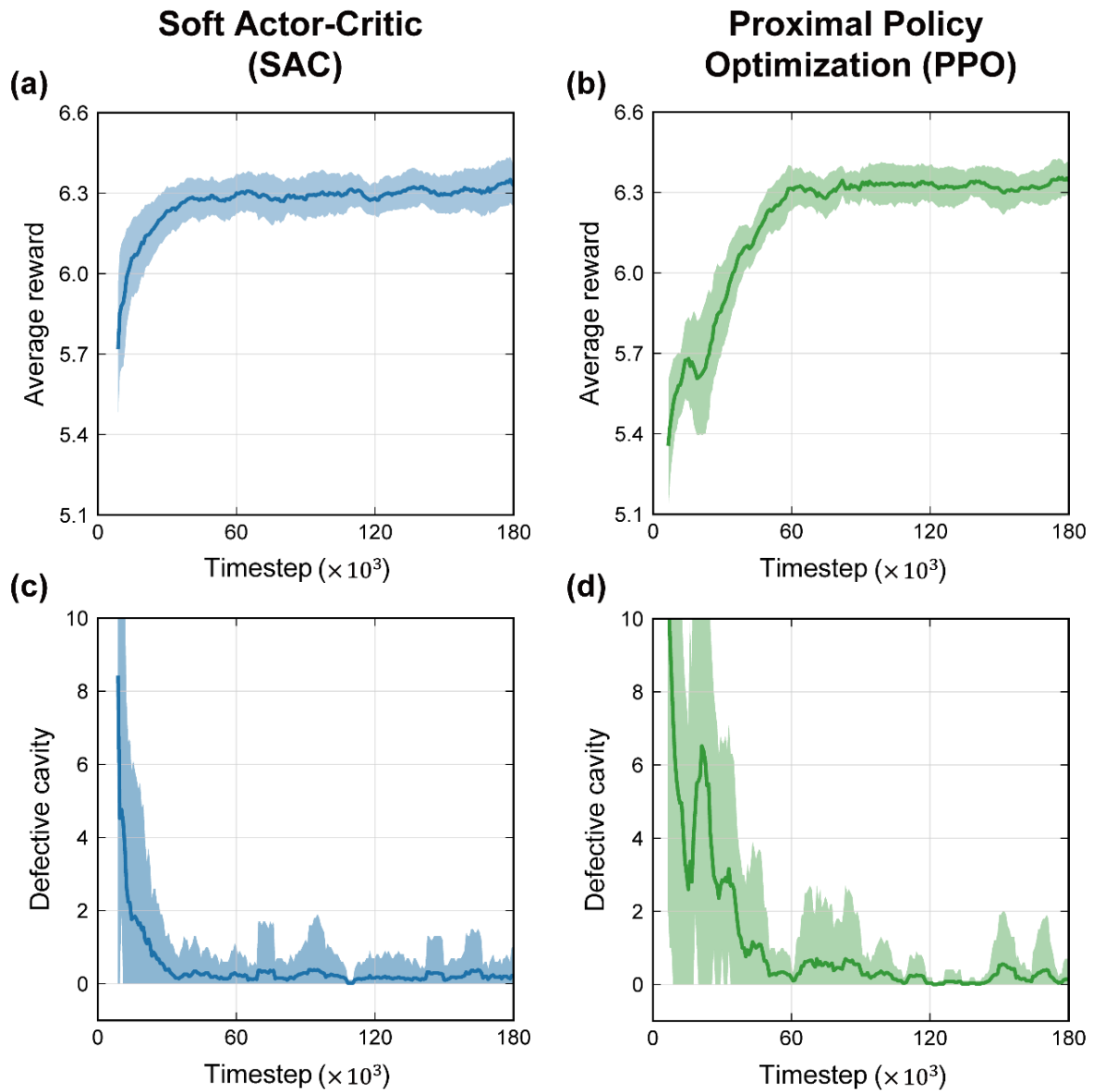

**Fig. 4**. Training curves of the average reward for models trained with (a) SAC and (b) PPO. Number of defective cavities during training of models using (c) SAC and (d) PPO. All curves are smoothed over 50 episodes using a moving average.

**Figs. 5a** and **5b** show the total accumulated profit obtained by testing the trained policy under a specific environmental variation scenario. In this scenario, the temperature and relative humidity variations correspond to the **Scenario 1** curves in **Figs. 6a** and **6b**, respectively, while the seasonal electricity price followed the spring price fluctuation pattern shown in **Fig. 2**. The trained policy makes a one-time adjustment to the process parameters in response to external environmental variations, aiming to maximize profit. In the test phase, the action was chosen deterministically as the mean of the policy distribution, in contrast to the training phase, where the action was stochastically sampled from the policy. When evaluating the average performance after 60,000 timesteps, where the policy had sufficiently converged, the average profits were 953.76 for SAC and 958.99 for PPO. The difference can be attributed to the inherent characteristics of the two algorithms, which lead to different timesteps required to reach the optimal solution with SAC requiring more timesteps than PPO.

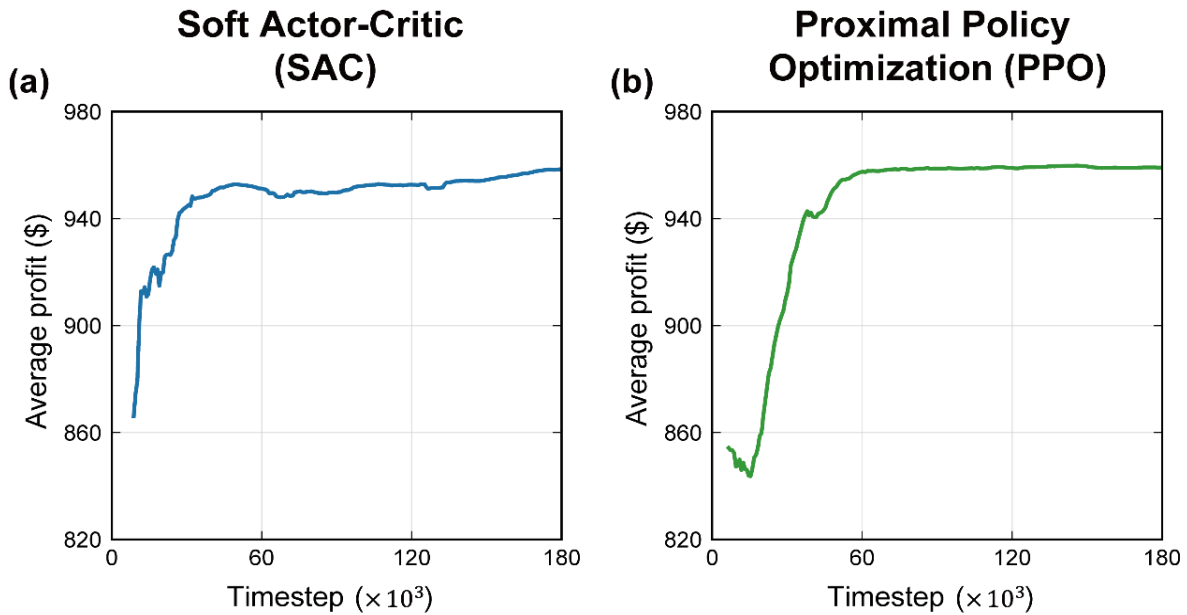

**Fig. 5**. Total profit over a single episode evaluated on a predefined scenario using policies trained with (a) SAC and (b) PPO. The scenario corresponds to "Scenario 1" in **Fig. 6**, which assumes seasonal electricity pricing based on spring/fall rates. Profit is evaluated using a deterministic version of the policy, where the mean action is selected instead of sampling from the stochastic distribution.

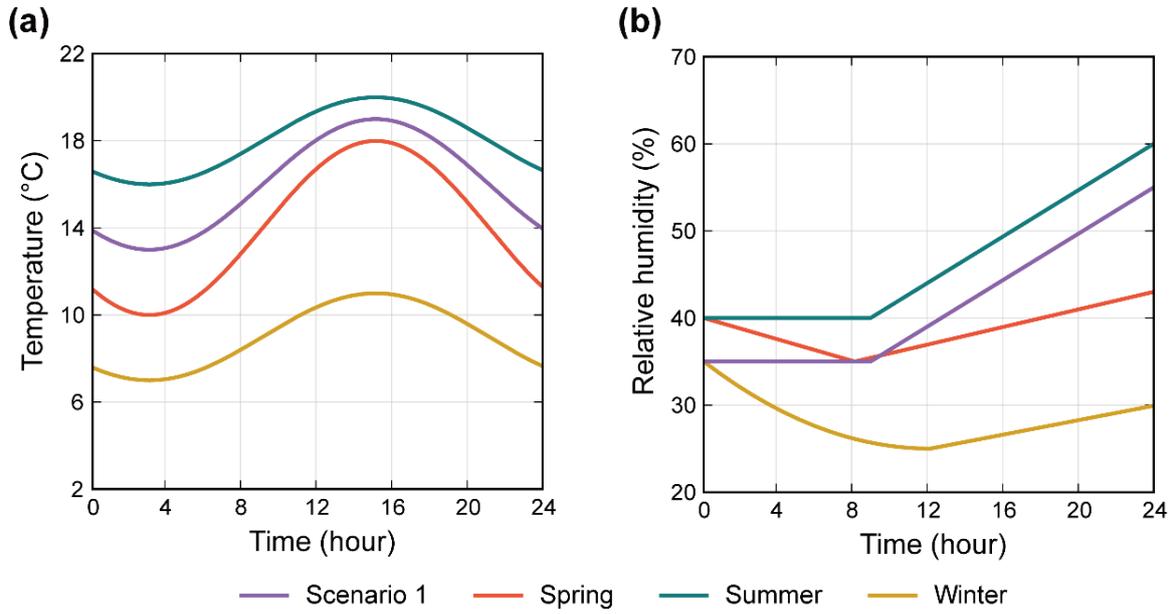

**Fig. 6**. Seasonal variations in environmental conditions for injection molding processes. (a) Temperature profiles for spring, summer, and winter. (b) Relative humidity profiles for spring, summer, and winter.

### 3.1.2 Comparison of the developed DRL models under specific environmental conditions

The performance of the DRL models trained in **Section 3.1.1** was evaluated to determine whether they could adjust process parameters from diverse initial process parameter settings to maximize profit. To this end, 9 initial process parameter cases were selected under a fixed environmental condition of 14°C temperature and 45% relative humidity for both the machine and factory with spring off-peak electricity pricing, as listed in **Table 8**. Six cases produced nondefective cavities with varying profit levels, whereas three resulted in defective cavities. Each initial process parameter case underwent a sequence of 10 consecutive adjustments, with each adjustment corresponding to one time-step. A deterministic policy was followed by taking the means of the policy distribution at every step, ensuring that identical inputs consistently yield identical actions. Parameters were updated cumulatively without reinitialization, so that the result of each adjustment served as the next state, forming a continuous trajectory of parameter evolution.

**Table 8.** Initial process parameter cases and corresponding profit and defective cavity count under spring off-peak conditions (temperature = 14°C, relative humidity = 45%)

| Process Parameter | Case | | | | | | | | |
|---|---|---|---|---|---|---|---|---|---|
| | 1 | 2 | 3 | 4 | 5 | 6 | 7 | 8 | 9 |
| Injection speed 1 | 32.5 | 38.8 | 26.2 | 36.0 | 44.4 | 30.5 | 37.6 | 44.3 | 38.2 |
| Injection speed 2 | 32.5 | 38.8 | 26.2 | 21.7 | 23.2 | 43.1 | 32.2 | 31.3 | 34.3 |
| Injection speed 3 | 20 | 25 | 15 | 12.8 | 12.8 | 25.6 | 25.4 | 12.0 | 12.4 |
| Injection pressure 1 | 130 | 135 | 125 | 131.5 | 136.2 | 120.7 | 121.14 | 133.5 | 124.7 |
| Injection pressure 2 | 130 | 135 | 125 | 133.4 | 127.1 | 120.8 | 132.17 | 122.5 | 125.0 |
| Injection pressure 3 | 140 | 145 | 135 | 132.2 | 144.3 | 133.0 | 140.36 | 134.7 | 137.8 |
| Injection position 1 | 46 | 47 | 45 | 45.6 | 45.2 | 45.4 | 46.0 | 47.4 | 44.4 |
| Injection position 2 | 38 | 41 | 35 | 40.0 | 34.7 | 40.7 | 40.3 | 39.4 | 41.1 |
| Injection position 3 | 30 | 31 | 29 | 30.7 | 29.6 | 30.5 | 28.3 | 31.3 | 29.9 |
| Hold time | 1.2 | 1.8 | 0.6 | 0.36 | 2.30 | 2.23 | 0.13 | 0.03 | 0.10 |
| Profit | 6.224 | 6.021 | 6.383 | 6.674 | 6.082 | 6.548 | 3.356 | 0.881 | -2.323 |
| Defective cavity count | 0 | 0 | 0 | 0 | 0 | 0 | 1 | 2 | 3 |

**Table 9.** Performance evaluation of SAC-base model: Profit variation by step across 9 initial process parameter cases under spring off-peak conditions (temperature = 14°C, relative humidity = 45%)

| Step | Profit by case | | | | | | | | |
|---|---|---|---|---|---|---|---|---|---|
| | 1 | 2 | 3 | 4 | 5 | 6 | 7 | 8 | 9 |
| 0(initial parameter) | 6.224 | 6.021 | 6.383 | 6.674 | 6.082 | 6.548 | 3.356 | 0.881 | -2.323 |
| 1 | 6.639 | 6.101 | 6.717 | 6.656 | 6.154 | 6.654 | 6.376 | 6.730 | 6.680 |
| 2 | 6.737 | 6.102 | 6.746 | 6.712 | 6.668 | 6.773 | 6.762 | 6.733 | 6.722 |
| 3 | 6.763 | 6.104 | 6.776 | 6.727 | 6.717 | 6.799 | 6.768 | 6.742 | 6.730 |
| 4 | 6.768 | 6.295 | 6.788 | 6.752 | 6.750 | 6.800 | 6.799 | 6.753 | 6.788 |
| 5 | 6.774 | 6.691 | 6.799 | 6.788 | 6.752 | 6.800 | 6.800 | 6.788 | 6.799 |
| 6 | 6.799 | 6.747 | 6.800 | 6.799 | 6.788 | 6.800 | 6.800 | 6.791 | 6.799 |
| 7 | 6.799 | 6.774 | 6.800 | 6.800 | 6.799 | 6.800 | 6.800 | 6.800 | 6.800 |
| 8 | 6.800 | 6.775 | 6.800 | 6.800 | 6.800 | 6.800 | 6.800 | 6.800 | 6.800 |
| 9 | 6.800 | 6.774 | 6.800 | 6.800 | 6.800 | 6.800 | 6.800 | 6.800 | 6.800 |
| 10 | 6.800 | 6.799 | 6.800 | 6.800 | 6.800 | 6.800 | 6.800 | 6.800 | 6.800 |
| Maximum Profit | 6.800 | 6.799 | 6.800 | 6.800 | 6.800 | 6.800 | 6.800 | 6.800 | 6.800 |

**Table 10.** Performance evaluation of PPO-based model: Profit variation by step across 9 initial process parameter cases under spring off-peak conditions (temperature = 14°C, relative humidity = 45%)

| Step | Profit by case | | | | | | | | |
|---|---|---|---|---|---|---|---|---|---|
| | 1 | 2 | 3 | 4 | 5 | 6 | 7 | 8 | 9 |
| 0(initial parameter) | 6.224 | 6.021 | 6.383 | 6.674 | 6.082 | 6.548 | 3.356 | 0.881 | -2.323 |
| 1 | 6.758 | 6.396 | 6.698 | 6.718 | 6.364 | 6.587 | 6.481 | 6.726 | 6.728 |
| 2 | 6.773 | 6.777 | 6.793 | 6.797 | 6.772 | 6.769 | 6.791 | 6.796 | 6.762 |
| 3 | 6.799 | 6.773 | 6.794 | 6.793 | 6.773 | 6.772 | 6.795 | 6.793 | 6.769 |
| 4 | 6.799 | 6.799 | 6.795 | 6.795 | 6.799 | 6.799 | 6.795 | 6.795 | 6.795 |
| 5 | 6.799 | 6.799 | 6.795 | 6.795 | 6.799 | 6.799 | 6.795 | 6.799 | 6.799 |
| 6 | 6.799 | 6.799 | 6.795 | 6.799 | 6.799 | 6.799 | 6.799 | 6.799 | 6.799 |
| 7 | 6.799 | 6.799 | 6.799 | 6.799 | 6.799 | 6.799 | 6.799 | 6.799 | 6.799 |
| 8 | 6.799 | 6.799 | 6.799 | 6.799 | 6.799 | 6.799 | 6.799 | 6.799 | 6.799 |
| 9 | 6.799 | 6.799 | 6.799 | 6.799 | 6.799 | 6.799 | 6.799 | 6.799 | 6.799 |
| 10 | 6.799 | 6.799 | 6.799 | 6.799 | 6.799 | 6.799 | 6.799 | 6.799 | 6.799 |
| Maximum Profit | 6.799 | 6.799 | 6.799 | 6.799 | 6.799 | 6.799 | 6.799 | 6.799 | 6.799 |

Both SAC- based and PPO-based models converged to profit-maximizing process parameters within 10 adjustment steps, regardless of the initial process parameter settings, and the respective results are presented in **Tables 9** and **10**. SAC-based model reached an average profit of 6.800 in 10 steps, whereas PPO-based model achieved 6.799 in 7 steps. The changes in average profit over adjustment steps for SAC-based and PPO-based model are illustrated in **Fig. 7**. In **Section 3.1.1**, PPO-based model yielded higher profit than SAC-based model when evaluated on the virtual scenario with a single adjustment from the DRL agent, which can be attributed to the characteristic of PPO to converge in fewer steps. Specifically, the mean first-step profit across all nine initial settings in **Tables 9 and 10** was 6.523 for the SAC-based model and 6.606 for the PPO-based model, which is consistent with the results presented in the previous section.

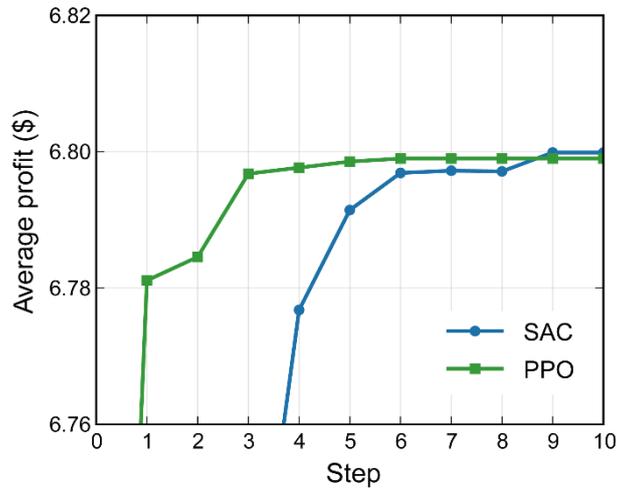

**Fig. 7**. Average profit by step for DRL models trained with SAC and PPO.

The results obtained using the DRL models were compared with those produced by a GA method utilizing the same surrogate models. GA is a population-based optimization method inspired by natural selection and genetics principles. It is one of the most widely adopted static approaches for solving complex global optimization problems and has been applied to injection molding studies [13,67]. This study used GA to search for process parameters that maximize the same profit function defined in **Eq. 1**, consistent with the DRL models. The optimization was performed under identical environmental conditions, including 14°C temperature, 45% relative humidity, and the spring off-peak electricity price. Using the DEAP library, we implemented the GA with simulated binary crossover and polynomial mutation, which are widely used in continuous, bounded optimization problems [68].Parent selection was performed using tournament selection, and the population size and number of generations were set to 40 and 25, respectively. The detailed hyperparameter settings for the GA are provided in **Section 2** of the **Supplementary Materials**.

**Fig. 8** shows the change in average profit for 9 initial populations over 25 generations. The results are compared with the profits obtained from the optimal parameters identified by SAC-based and PPO-based model. The GA converged to an average profit of 6.799 after 20 generations (800 profit evaluations) with profit ranging from 6.784 to 6.805 and a standard deviation of 0.00613.

For the DRL models, SAC-based and PPO-based model achieved profits of 6.799 and 6.800, respectively. Despite being trained in a stochastic manner, the model produced consistent results across trials in the deployment phase due to the use of a deterministic policy. This highlights the high robustness of the DRL model under the given environmental conditions, while still achieving performance comparable to the GA.

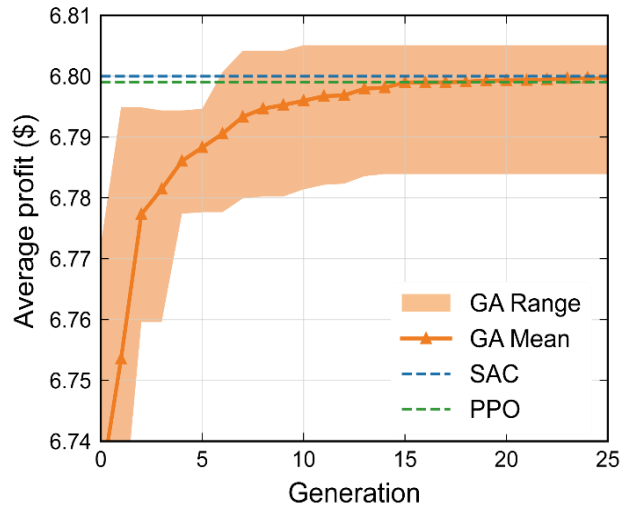

**Fig. 8**. Evolution of the average best profit during genetic algorithm optimization, performed with 9 random initial populations under spring off-peak conditions (temperature = 14°C, relative humidity = 45%). The shaded region represents the range of fitness variation across populations. Results are compared with the average profits achieved by SAC-based and PPO-based model after 10 and 7 adjustment steps, respectively.

**Table 11.** Comparison of DRL models and GA

| Method | Average profit | Standard deviation | Computational time | Training time |
| --- | --- | --- | --- | --- |
| DRL – SAC | 6.800 | – | 0.421s (10 steps) | 206 min |
| DRL – PPO | 6.799 | – | 0.287s (7 steps) | 182 min |
| GA | 6.799 | 0.00613 | 21.201s (20 gens) | – |

**Table 11** presents the optimization and training times as well as the average profit and standard deviation for the two DRL models and the GA method. As shown in the previous results, SAC-based and PPO-based model required 10 and 7 profit evaluations respectively to reach convergence, while the GA required 20 generations, corresponding to a total of 800 evaluations. The total training times were 206 minutes for SAC-based model and 182 minutes for PPO-based model, while the

GA required no prior training. However, it is important to note that the training process for the DRL models can be conducted offline, before deployment, and does not need to occur within the production environment. As such, the training time is not a critical constraint in practical applications. In contrast, the time required for online optimization is a key constraint in injection molding processes, where optimization must be completed within the cycle time. In this regard, the DRL-based approaches demonstrated a significant advantage by performing optimization tens of times faster than the GA, achieving inference times of 0.421 seconds for SAC-based model and 0.287 seconds for PPO-based model, compared to 21.201 seconds for the GA. Furthermore, the GA searches for optimal process parameters under fixed environmental conditions, whereas DRL incorporates environmental variables into the state representation, enabling dynamic optimization that can adapt to changing conditions. To properly assess the effectiveness of the trained DRL models, it is important to verify their ability to perform reliably under fixed environmental conditions and across a diverse range of scenarios, which is evaluated in **Section 3.1.3**.

### 3.1.3 Evaluation of the developed DRL models under varying environmental conditions

To assess whether the trained DRL models can track optimal process parameters under dynamic conditions, we performed a 24-hour virtual deployment in three seasonally representative scenarios including spring, summer, and winter. In South Korea, spring and autumn electricity pricing and environmental characteristics are nearly identical. Accordingly, spring, summer, and winter were selected as representative seasons for evaluating the effects of seasonal variations. It is important to note that the scenarios representing seasonal variations can be adapted according to country-specific environmental characteristics and conditions. Each scenario followed the corresponding temperature–humidity trajectory shown in **Fig. 6**, under the simplifying assumption that the machine and factory shared identical seasonal conditions at every time step. These profiles were selected to cover various seasonal conditions while remaining within the experimental bounds listed in **Table 1**, ensuring that the trained policy could adapt to practical operating environments.

The cumulative profit defined in **Eq. 1** over a 24-hour operation of a single injection molding machine was compared with DRL models and GA method under the three seasonal scenarios. The machine was assumed to operate continuously for 24 hours, with production occurring immediately after each cycle. Process parameters were optimized for all methods based on the temperature and relative humidity measured at the beginning of each cycle. For the DRL models, the final performance was consistent across different initial process parameters (**Section 3.1.2**); thus, the initial condition for optimization was set to the midpoint (50%) of the range listed in **Table 1**, corresponding to a normalized state $s_0 = 0$. The number of steps required for convergence was determined to be 10 steps for SAC-based model and 7 steps for PPO-based model, based on the results in **Tables 9** and **10**, and the process parameters obtained at these steps were adopted as the optimized parameters. The same hyperparameter settings used in **Section 3.1.2** were applied for the GA, and optimization was conducted at each cycle based on the updated environmental conditions.

**Figs. 9a, 9b,** and **9c** show the cumulative profit changes over 24 hours for the DRL models trained with SAC and PPO, and for the GA method, under three seasonal scenarios. For all methods, the cumulative profit consistently followed the order of spring, winter, and summer, corresponding to the decreasing electricity prices shown in **Fig. 2**. **Table 12** summarizes the total profit, the number of cavities produced over 24 hours, and the computational time measured for the spring scenario. SAC-based model achieved profits of $958.88, $915.63, and $930.85 in the spring, summer, and winter scenarios, respectively, while PPO-based model achieved profits of $958.33, $914.68, and $929.85 under the same scenarios. Meanwhile, SAC-based model produced 8,644, 8,744, and 8,824 cavities, whereas PPO-based model produced 8,640, 8,736, and 8,816 cavities

under the same scenarios. The GA recorded profits of $959.69, $915.87, and $932.66 and produced 8,652, 8,748, and 8,844 cavities in the spring, summer, and winter scenarios, respectively. Although the GA achieved the highest profits and production volumes across all scenarios, the DRL models exhibited highly comparable performance in both economic and operational terms, demonstrating their practical competitiveness under varying environmental conditions.

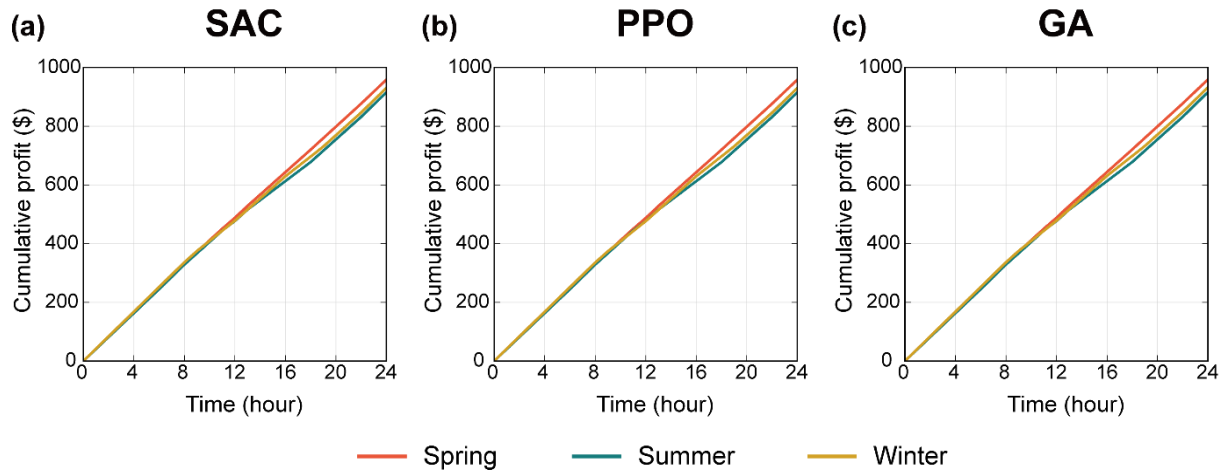

**Fig. 9**. Cumulative profit over a 24-hour period under varying environmental conditions is presented for 2 DRL models trained using (a) SAC and (b) PPO, and (c) the GA method. All methods were deployed under dynamic environmental conditions, incorporating the temperature and relative humidity profiles shown in **Fig. 6** and the electricity price variations illustrated in **Fig. 2**. Across all approaches, the seasonal ranking of cumulative profit remains consistent (spring > winter > summer).

While the DRL models and the GA method exhibited highly comparable performance levels, their computational efficiency differed markedly. In this study, computational time refers to inference time for the trained DRL models and full optimization time for the GA method. Inference time refers to the time required for a trained DRL model to produce process parameter adjusetment decision via multiple forward passes of the policy network. The SAC-based and PPO-based models required only 15.5 minutes and 10.4 minutes, respectively, primarily due to differences in the number of steps required to reach convergence. In contrast, the GA method required a substantially longer time of 781.0 minutes. Importantly, once trained to accommodate dynamic environmental variations, the DRL models can infer optimal process parameters through a single forward pass of the policy network, incurring minimal computational overhead during operation. Conversely, the GA method must perform a complete optimization routine whenever environmental conditions change, resulting in significant and persistent computational costs during real-time deployment.

**Table 12** Comparison of total profit and number of cavities produced (in parentheses) over 24 hours by season for DRL models (trained using SAC and PPO) and the GA method. Computational time is measured based on the spring scenario.

|  | SAC | PPO | GA |
| --- | --- | --- | --- |
| Spring | $958.88 (8,644) | $958.33 (8,640) | $959.69 (8,652) |
| Summer | $915.63 (8,744) | $914.68 (8,736) | $915.87 (8,748) |
| Winter | $930.85 (8,824) | $929.85 (8,816) | $932.66 (8,844) |
| Computational time | 15.5 min | 10.4 min | 781.0 min |

### 3.2 Comparison of the DRL models under different adjustment step size settings

The impact of different adjustment step sizes, as presented in **Table 5,** on the performance of the DRL models was systematically analyzed. **Section 3.1.3** evaluated the models using a large adjustment step size across 3 seasonal scenarios. This section examines a smaller adjustment step size for comparative analysis. Two cases were considered under the small adjustment condition: (1) maintaining the number of steps identical to the large adjustment case (10 steps for SAC, 7 steps for PPO), and (2) doubling the number of steps (20 steps for SAC, 14 steps for PPO).

**Tables 13** and **14** provide quantitative comparisons of cumulative profit and the number of cavities produced over 24 hours for SAC-based and PPO-based models under different step size and step count conditions. When the smaller adjustment step size was employed with the same number of steps (10 for SAC, 7 for PPO), the SAC-based model experienced a decrease in both profit and production, particularly during summer and winter. Similarly, the PPO-based model showed decreased performance in spring and summer scenarios, although a slight profit increase occurred in winter. Increasing the number of steps to 20 for SAC and 14 for PPO improved performance for both models. Specifically, SAC achieved near-equivalent performance to the large adjustment scenario, whereas PPO showed improved but slightly lower outcomes. Computational times for both models increased proportionally with the number of steps. **Fig.10** shows the total profit the total profit and number of cavities over 24 hours for SAC- and PPO-based models across three seasonal senarios and varying step conditions, summarizing the results presented in **Tables 12, 13,** and **14**.

**Table 13.** Comparison of total profit and number of cavities produced (in parentheses) over 24 hours for SAC-based model operated with different step sizes and numbers of steps. Computational time is measured based on the spring scenario.

|  | SAC (large step size, 10 steps) | SAC (small step size, 10 steps) | SAC (small step size, 20 steps) |
|---|---|---|---|
| Spring | $958.88 (8,644) | $958.79(8,644) | $958.79(8,644) |
| Summer | $915.63 (8,744) | $912.91(8,720) | $914.29 (8,732) |
| Winter | $930.85 (8,824) | $928.64 (8,804) | $929.09 (8,808) |
| Computational time | 15.5 min | 15.3 min | 30.5 min |

**Table 14.** Comparison of total profit and number of cavities produced (in parentheses) over 24 hours for PPO-based model operated with different step sizes and numbers of steps. Computational time is measured based on the spring scenario.

|  | PPO (large step size, 7 steps) | PPO (small step size, 7 steps) | PPO (small step size, 14 steps) |
|---|---|---|---|
| Spring | $958.33 (8,640) | $952.81 (8,592) | $956.04(8,620) |
| Summer | $914.68 (8,736) | $909.70 (8,688) | $911.54 (8,704) |
| Winter | $929.85 (8,816) | $930.76 (8,824) | $930.76 (8,824) |
| Computational time | 10.4 min | 10.3 min | 20.5 min |

The observed slight profit increase for the PPO-based model in the winter scenario under the smaller adjustment setting (7 steps) is not considered decisive. Evaluating DRL models must emphasize their capability to consistently identify optimal parameters across diverse environmental conditions, rather than isolated improvements in specific scenarios. Moreover, while doubling the step count improved performance, this enhancement was accompanied by significantly increased computational time and operational costs. Thus, a smaller adjustment step size demands greater computational resources for convergence without guaranteeing a substantial or consistent advantage. Consequently, a larger adjustment step size emerges as the preferable choice, offering a robust and computationally efficient strategy for real-time process optimization.

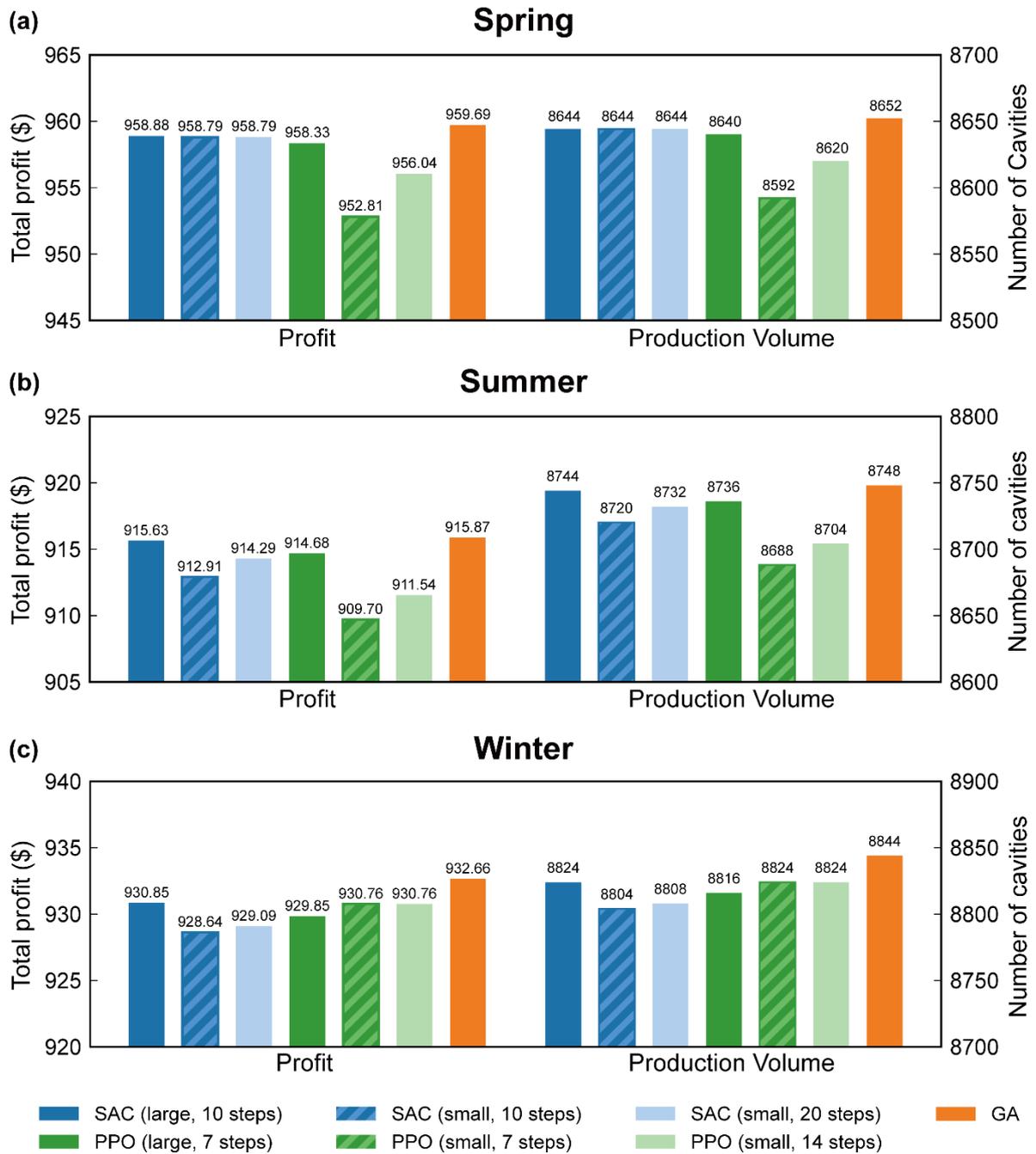

**Fig. 10**. Comparison of total profit and number of cavities produced over 24 hours for SAC-based and PPO-based model operated with different step sizes and numbers of steps under three seasonal scenarios (spring, summer, and winter)

## 4  DISCUSSION

### 4.1  Performance Discussion of DRL-based decision-making models

We developed decision-making models based on two DRL algorithms, SAC and PPO, to identify optimal process parameters under dynamic environmental conditions reflecting temporal and seasonal variations in temperature, relative humidity, and electricity prices. Under the large adjustment step size condition, comparisons across three virtual seasonal scenarios showed that SAC-based model achieved slightly better cumulative profit and production volume compared to PPO-based model. However, due to the characteristics of its maximum entropy framework, SAC maintains higher policy stochasticity to explore a broader action space, requiring more steps to converge than PPO.

Under the small adjustment step size condition, optimization required a greater number of steps; nevertheless, SAC-based model was able to maintain performance comparable to that under the large step size setting, while PPO-based model showed a stronger tendency to converge to a locally optimal policy. The inefficiency and extended convergence time resulting from smaller adjustment steps have also been observed in previous research that applied DDPG for optimizing injection molding processes [44]. Off-policy algorithms such as SAC and DDPG can reuse previously collected data, enabling broader exploration of the state-action space [35,65]. In contrast, the on-policy PPO algorithm relies solely on newly collected trajectories, which limits data diversity and narrows the exploration range [34]. This algorithmic difference was reflected in our small adjustment step size experiments, as SAC could approach near-global optima, whereas PPO more frequently settled into locally optimal solutions.

The optimization performance of the developed DRL-based models was evaluated against a static method, GA, under diverse environmental conditions. The SAC-based model could achieve results comparable to the GA, with profit differences of $0.81, $0.24, and $1.81 over 24 hours for the spring, summer, and winter scenarios, respectively. In the deployment phase, SAC-based model

required 15.5 minutes to optimize a single scenario, while the GA took 781 minutes, demonstrating that DRL approach is much more efficient in terms of optimization cost. The faster inference speed of the DRL models was achieved by training them not only with process parameters but also with environmental variables and electricity prices included in the state representation, allowing them to adapt to dynamic environmental changes. In contrast, the GA optimizes process parameters under fixed conditions and must be rerun whenever the environment changes. While this can yield more precise results in static settings, it is inefficient for real-time applications under continuously varying conditions.

This difference becomes even more critical as the number of process parameters increases since the optimization time of the GA scales with the input dimensionality [69], whereas the inference time of the DRL-based models remains effectively constant. In particular, the previous study identified 10 key process parameters among 43 candidates based on their practical adjustability by field experts, and this study builds upon that selection. If additional process paramters must be considered, the computational cost of the GA rises sharply, making real-time deployment even more challenging. In contrast, although the DRL models becomes more complicated to train with a greater number of variables, it can still achieve rapid inference speed once training is completed, making it more suitable for real-time adaptation under continuously changing conditions.

### 4.2 Expectation of DRL-based decision-making models

The proposed DRL-based decision-making models, utilizing SAC and PPO, dynamically adapt to seasonal and time-dependent fluctuations while optimizing process parameters to enhance production quality and profitability. Leveraging DRL supports adaptive and profitable production, enabling real-time decision-making in evolving manufacturing environments.

A key strength of these models is their ability to generalize across diverse environmental conditions. Once trained, the DRL models can efficiently determine optimal process parameters that ensure consistent profitability, even when fluctuations in electricity pricing, environmental factors, and production constraints occur. Unlike static optimization methods, DRL-based dynamic optimization enables real-time adaptation, maintaining both cost-efficiency and production performance. Comparative experiments with the GA, a widely used global optimization method, confirm that the DRL-based models achieve comparable profitability while reducing optimization time by up to 135 times. This computational efficiency is a technical advantage and a critical factor for practical deployment in real-world manufacturing environments.

Field experts apply a recommended process parameter setting in actual manufacturing environments, especially in injection molding. These experts need enough time to review the suggested parameters, assess their plausibility, and input them into the molding machine before the next cycle begins. Given that the target product's cycle time in this study is approximately 39 seconds, the GA's optimization time of 21 seconds alone consumes more than half of the available cycle time. This severely limits the remaining time available for expert verification and manual setup, making GA impractical for real-time field deployment.

Although the GA may achieve slightly higher profitability under some seasonal settings, with gains ranging from $0.86 to $2.81 per 24-hour operation compared to SAC-based or PPO-based models, realizing such improvements would require significantly higher computational specifications, which in turn leads to increased investment capital. In the injection molding industry, which typically operates under low-margin, cost-sensitive conditions, investing in high-performance computing infrastructure solely to support GA-based optimization is economically inefficient and unrealistic. Factories in this sector are often reluctant to invest in computational upgrades, prioritizing low operational costs over marginal gains in theoretical optimization performance. In contrast, the DRL models offer a cost-effective and scalable solution. Their rapid inference time aligns well with the actual production cycle, allowing field experts sufficient time for evaluation and implementation without disrupting production flow.

Importantly, the generality of the proposed models is largely driven by their profit function formulation. Since the profit function can be customized to reflect the economic goals and constraints of different manufacturing contexts, the DRL-based dynamic optimization framework are not limited to plastic injection molding. Instead, they provide a scalable and transferable solution for real-time process optimization in other industries, such as semiconductor fabrication, metal forming, and food processing, where dynamic operating conditions and cost variables influence profitability. By combining data-driven insights with reinforcement learning, this study presents a flexible decision-making framework capable of adapting to a wide range of production systems that require rapid, cost-effective, and stable operation.

### 4.3 Limitation of DRL-based decision-making models

One of the principal limitations of this research is the dependence of DRL performance on the accuracy of the surrogate model, which serves as the offline training environment for the DRL agent. The success of DRL-based optimization critically hinges on the surrogate model's ability to approximate the real injection molding process faithfully. Suppose the surrogate model lacks sufficient precision due to measurement errors, environmental uncertainties, or inherent modeling limitations. In that case, the DRL agent may learn suboptimal policies that fail to generalize to actual production environments. Significant discrepancies between the surrogate model and the real-world process can result in policies that perform poorly when deployed, leading to unexpected declines in product quality or profitability and ultimately compromising the effectiveness of the DRL-based optimization. If the surrogate model lacks sufficient training data, its predictive accuracy remains limited, ultimately affecting the optimization quality of the DRL-based model. Thus, securing extensive datasets is crucial for improving overall system performance. Ensuring high surrogate model fidelity necessitates using a sufficiently large and representative dataset for training. However, in industrial environments, data collection is often constrained by restricted access to production systems, variability in operating conditions, and high acquisition costs. Training the surrogate model on limited or biased datasets can degrade its predictive accuracy, thereby diminishing the optimization quality and robustness of the DRL model. Therefore, securing extensive and diverse datasets is critical to enhancing surrogate model fidelity, enabling the DRL agent to develop reliable and transferable policies to real production systems.

Another limitation arises as the number of process parameters increases or interdependencies among parameters become more complex. Under such conditions, the single-agent DRL approach employed in this study may struggle to efficiently explore the solution space and identify globally optimal process parameters. The increased complexity can cause the agent to converge to local optima, thus hindering the achievement of the most effective production settings. To overcome this challenge, multi-agent DRL algorithms should be considered. The multi-agent algorithms would allow individual agents to specialize in optimizing subsets of process parameters while collaborating to achieve a globally optimized solution. This approach could enhance adaptability and scalability in highly complex manufacturing environments, where single-agent DRL algorithms may encounter performance limitations.

### 4.4 Future work

While the proposed DRL-based decision-making models have demonstrated their effectiveness in dynamically adapting to changing manufacturing environments, further advancements are necessary to address certain limitations and enhance practical applicability. One key area for future research is improving the surrogate model by expanding the available dataset. Since the accuracy of the surrogate model directly impacts the performance of the DRL model, securing a sufficient and diverse dataset is essential. However, operational limitations often constrain industrial data collection, which necessitates alternative approaches to improve model robustness. A promising direction is the generation of synthetic data to supplement real-world datasets. This can be achieved by leveraging multi-fidelity simulation techniques, which integrate high- and low-fidelity models to create realistic training environments for the DRL agent. Additionally, transfer learning and few-shot learning can be explored to enhance the model's adaptability when only limited real-world data are available. By incorporating these techniques, future studies can alleviate data scarcity issues and improve the overall accuracy and reliability of the surrogate model.

Furthermore, as the number of process parameters increases or the interdependencies among variables become more complex, single-agent DRL algorithms may struggle to optimize all parameters efficiently. In such scenarios, multi-agent DRL algorithms should be considered to overcome local optima and improve scalability. A multi-agent framework allows individual agents to optimize specific process variables while collaborating to achieve a globally optimal solution. This approach can significantly enhance the flexibility and robustness of DRL-based optimization, making it more suitable for large-scale and complex manufacturing processes.

By refining the surrogate model and investigating multi-agent reinforcement learning, future research can further improve the adaptability, efficiency, and scalability of DRL-based optimization for injection molding and broader industrial applications. These advancements will support the development of more intelligent, data-efficient, and real-time adaptive manufacturing systems, ensuring higher reliability in highly dynamic production environments.

## 5  CONCLUSION

This study presented a Deep Reinforcement Learning (DRL)-based framework and models for optimizing injection molding process parameters with the dual objectives of improving product quality and maximizing profitability. The proposed framework addressed a critical limitation of conventional optimization approaches by formulating a profit-driven objective function that comprehensively integrates material costs, mold maintenance costs, electricity consumption, and real-time electricity pricing. To enable efficient offline training, surrogate machine learning models were developed for quality classification and cycle time prediction, allowing DRL agents to be trained without direct interaction with physical production systems.

Experimental results demonstrated that the developed DRL-based decision-making models, utilizing SAC and PPO algorithms, effectively adapted process parameters to fluctuating environmental conditions. The models achieved profitability comparable to a genetic algorithm while significantly reducing optimization time, offering a clear computational advantage. The rapid inference capability enables parameter optimization within seconds, making the framework highly suitable for real-time deployment in manufacturing environments where timely and adaptive decision-making is essential.

The key contributions of this research are summarized as follows. First, a profit-driven optimization strategy was proposed that directly links operational decisions to economic outcomes, addressing a gap often overlooked in conventional quality-focused approaches. Second, a DRL-based real-time decision-making framework was developed, demonstrating stable and efficient optimization performance under dynamic environmental and economic conditions. Third, the proposed models exhibited robust adaptability across diverse seasonal scenarios, validating their generalizability in real-world manufacturing environments. Fourth, the framework was designed to be scalable and transferable to other industrial domains, highlighting its potential applicability beyond injection molding. Finally, by identifying limitations related to surrogate modeling and optimization complexity, this study lays the foundation for future research to improve system scalability through enhanced data-driven modeling and multi-agent reinforcement learning strategies.

Overall, this research underscores the transformative potential of DRL-based optimization in smart manufacturing systems. Future work will focus on expanding datasets to improve surrogate model fidelity and investigating multi-agent reinforcement learning approaches to enhance scalability, robustness, and adaptability in increasingly complex industrial environments.

**Acknowledgements:** This work was supported by the Technology Innovation Program (RS-2023-00284506) funded by the Ministry of Trade, Industry and Energy (MOTIE, Korea), the Technology Development Program (S3207585) funded by the Ministry of SMEs and Startups (MSS, Korea), the National Research Foundation of Korea (NRF) grant funded by the Ministry of Science and ICT (RS-2023-00222166), and a grant from the Ministry of Food and Drug Safety (RS-2023-00215667).

**Data Availability Statement:** The data used are not publicly available.

**Conflicts of Interest:** The authors declare no conflict of interest.